\documentclass{article}
\usepackage{spconf,amsmath,graphicx}
\usepackage{threeparttable}
\usepackage{epstopdf}
\usepackage{amsfonts}
\usepackage{subfigure}
\usepackage{color}
\usepackage{multirow, multicol, makecell}
\usepackage{setspace}

\usepackage{booktabs}
\usepackage{array}

\usepackage[colorlinks,linkcolor=black,anchorcolor=black,urlcolor=black,citecolor=black]{hyperref}
\usepackage{url}

\title{Enhancing Quality for VVC Compressed Videos by Jointly Exploiting Spatial Details and Temporal Structure}%
\name{Xiandong Meng$^{1}$, Xuan Deng$^{2}$, Shuyuan Zhu$^{2}$ and Bing Zeng$^{2}$}
\address{$^{1}$The Hong Kong University of Science and Technology\\
$^{2}$University of Electronic Science and Technology of China
}

\begin{document}
\maketitle
\begin{abstract}
In this paper, we propose a quality enhancement network of versatile video coding (VVC) compressed videos by jointly exploiting spatial details and temporal structure (SDTS). The proposed network consists of a temporal structure fusion subnet and a spatial detail enhancement subnet. The former subnet is used to estimate and compensate the temporal motion across frames, and the latter subnet is used to reduce the compression artifacts and enhance the reconstruction quality of compressed video. 
Experimental results demonstrate the effectiveness of our SDTS-based method. The code of our proposed method is available at \textit{\url{https://github.com/mengab/SDTS}}
\end{abstract}


\section{Introduction}
Versatile video coding (VVC) \cite{PCS1} achieves a higher compression performance compared with the High Efficiency Video Coding (HEVC) \cite{Sullivan1}. Similar to previous video coding standards, the VVC also employs a hybrid scheme which includes the block-based prediction and transform coding to compress videos. Due to the quantization of the transform coefficients in each small block, the artifacts, such as the blocking artifacts and the ringing effects, usually exist in the compressed videos, especially at the low bit-rate. Therefore, it is necessary to enhance the quality of compressed video.

In this work, we focus on the quality enhancement for the compressed video signals based on the latest convolutional neural network (CNN) method. The video quality enhancement may be regarded as the extension of the image quality enhancement in the temporal dimension. Such an extension introduces more prior information which can be used to potentially improve the quality of each individual frame. However, there still exist some challenges to utilize these information to construct an efficient CNN-based solution. 
First, removing compression artifacts from videos requires the understanding of not only the spatial context of the single frame but also the motion information across frames. 
Second, although it is possible to find missing content of the same scene or object in adjacent frames, the interference information will be introduced to the target frame if the adjacent frames are directly input to network as a reference. 
Third, due to the quality fluctuation across compressed video frames, it is very difficult to enhance all video frames with a single model.

\begin{figure}[t]
\centering
\includegraphics[width=7.0cm]{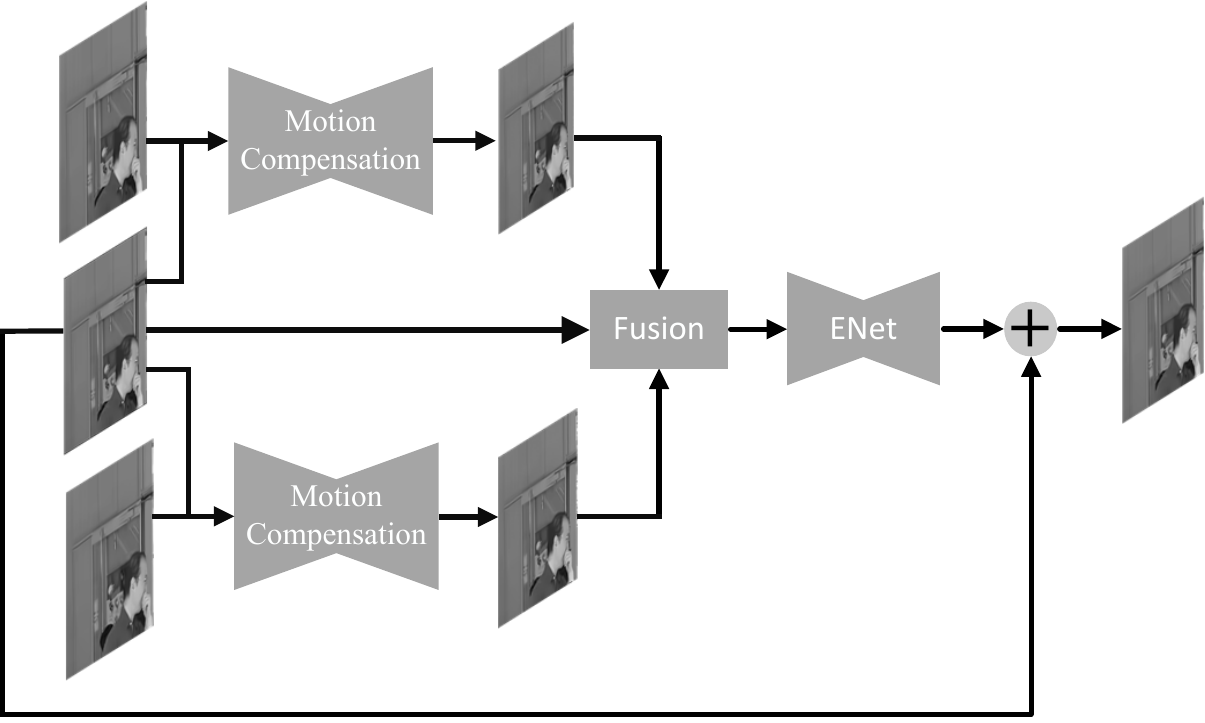}
\caption{The framework of our proposed SDTS-based method}
\end{figure}

We propose a novel end-to-end deep learning architecture in this work to tackle the above issues. The framework of the proposed network is shown in Fig. 1, which consists of a temporal fusion subnet and a spatial detail enhancement subnet. The first subnet is utilized to estimate and compensate the temporal motion across frames, and the second one is employed to reduce the compression artifacts. In addition, as pointed out in \cite{key1,key2,key4,Yang16} that the low-quality frames (LQFs) may be enhanced using the adjacent high-quality frames (HQFs), we also employ the adjacent HQFs  as a reference to enhance the low-quality frames. The experimental results demonstrate the performance of the SDTS-based method.

\section{Related Work}
Deep learning has been successfully applied to video super-resolution \cite{Caballero}, deblurring \cite{Unet3} and inpainting \cite{chuan1}, and can also be employed to enhance the quality of compressed image/video \cite{Yang16, Dai14, Dong9_15,Tai17, Wang15_17Chao, Wang12_16, meng1,Zhang13_17,partition}. Particularly, Dong \textit {et al.} \cite{Dong9_15} firstly proposed ARCNN to reduce the JPEG artifacts of images. Later on, DnCNN \cite{Zhang13_17} and MemNet \cite{Tai17} were proposed for image restoration, including the image quality enhancement.
For the quality enhancement of compressed video, VRCNN \cite{Dai14} was proposed as a variable-filter-size residue-learning network \cite{res} for the post-processing of HEVC intra coding. Wang \textit {et al.} \cite{Wang15_17Chao} developed a Deep CNN-based Auto Decoder (DCAD), which contains 10 CNN layers to reduce the distortion of compressed video. These methods were proposed based only on the prior information of a single frame,  so the enhancement performance is still limited. To tackle this problem, Yang \textit {et al.} \cite{Yang16} proposed a MFQE model with multi-frame input for quality enhancement of HEVC compressed video in which the information of neighboring key frames was considered. Meanwhile, Meng \textit {et al.} \cite {meng1} designed a multi-frame guided attention network by taking advantage of the intra-frame prior information and multi-frame information to enhance the quality of the HEVC compressed video. The experimental results of \cite{Yang16} and \cite {meng1} have demonstrated that utilizing the multi-frame information to build up the network for video quality enhancement can achieve excellent performance.

\section{Our Proposed method}\label{sec:format}
The proposed network consists of a temporal fusion subnet and a spatial detail enhancement subnet. Particularly, the temporal fusion subnet has two key modules, i.e., the motion compensation (MC) module and the fusion module. In this section, we focus on the design of  these three modules.
\subsection{MC Module}
The multi-frame video processing networks are normally built upon the fact that different observations of the same object or scene are probably available across frames of a video. As a result, content or scene, which are lost due to certain processing on the target frame, may be found in adjacent frames. Therefore, an intuitive idea is to enhance the compression quality of target frame by directly inputting multiple frames to the network. However, due to inter-frame motion, the interference information may be introduced to the network, especially for those scenes with drastic motion. To tackle this problem, we firstly employ a subnet to estimate and compensate the temporal motion across frames. Then, the compensated adjacent frames are used to enhance the quality of target frame.

In \cite{Caballero}, Caballero \textit {et al.} proposed the spatial transformer motion compensation (STMC) for video super-resolution. The basic idea of STMC is to predict the optical flow of adjacent frames to current frame by multi-scale down-sampling network. Suppose $I_t$ and $I_{t+1}$ are two consecutive frames, the optical flow related to adjacent frame $I_{t+1}$, whose reference frame is $I_{t}$, is a function of motion parameter ${\theta _{\Delta ,t + 1}}$. This optical flow can be represented by two feature maps corresponding to displacements of the $x$ and $y$ dimensions, i.e., ${\Delta^x_{t + 1}}$ and ${\Delta^y_{t + 1}}$, as ${\Delta _{t + 1}} = \left( {{\Delta^x_{t + 1}},{\Delta^y_{t + 1}};{\theta _{\Delta ,t + 1}}} \right)$. Then, the compensated frame $I_{t+1}^\prime$ can be expressed as
\begin{equation}
I_{t+1}^\prime\left( {x,y} \right) = {\cal I}\left\{ {{I_{t + 1}}\left( {x + {\Delta^x _{t + 1}},y + {\Delta^y_{t + 1}}} \right)} \right\},
\end{equation}
where ${\cal I}$ denotes the bilinear interpolation. STMC consists of a coarse ($\times4$) and a fine ($\times2$) scale optical flow estimation.

\begin{figure}[t]
\centering
\includegraphics[width=8.2cm,height=4.3cm]{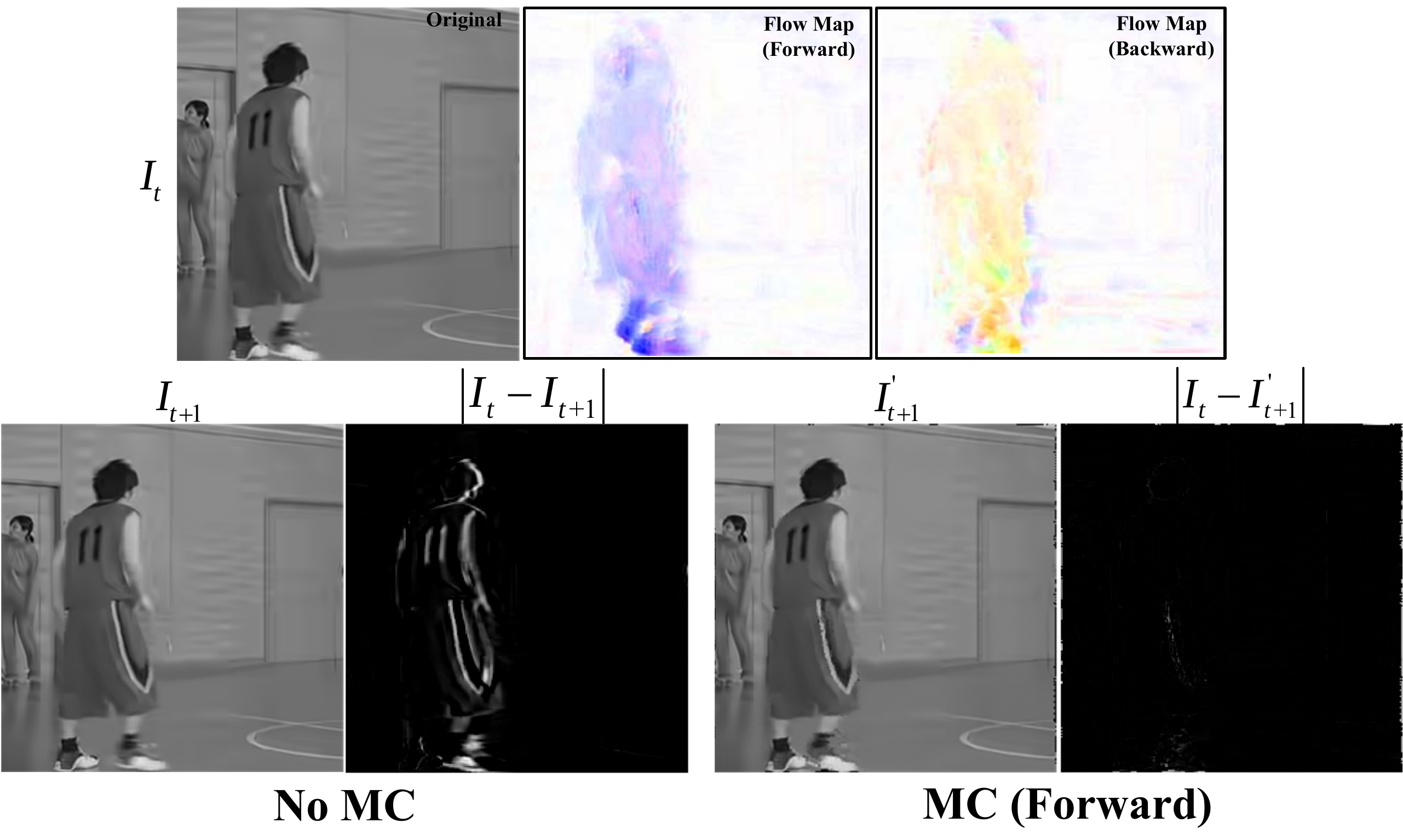}
\caption{Top: flow map estimated relating the original frame. Bottom: the consecutive frames without and with MC (No MC and MC).}
\vspace{-0.6em}
\end{figure}

\begin{figure*}[t]
\centering
\includegraphics[width=18.0cm] {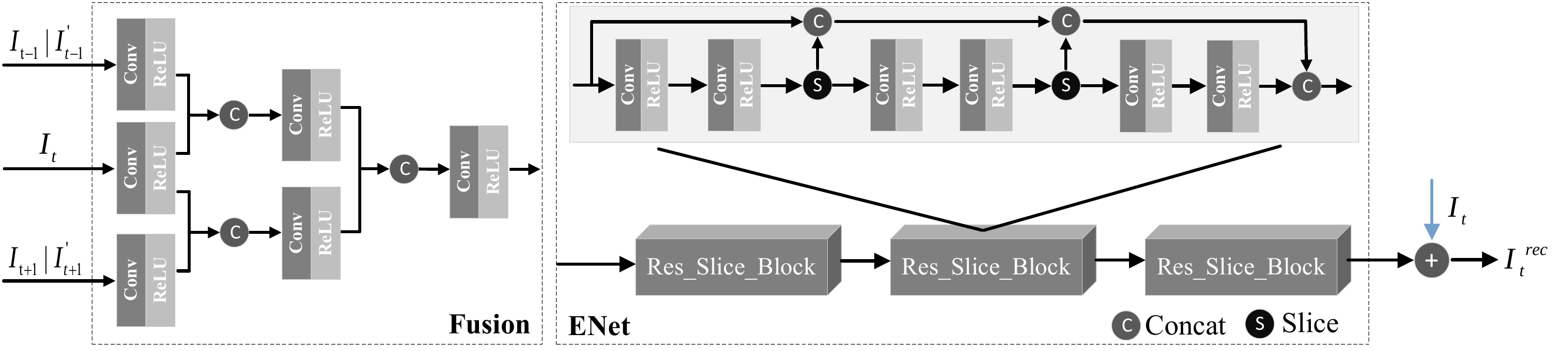}
\caption{The temporal fusion module and spatial detail enhancement subnet.} %
\end{figure*}

We make several modifications on STMC to adapt it to our proposed SDTS method.
First, we employ the coarse-to-fine ($\times4$ and $\times2$) flow estimation modules to handle large scale motion. Also, we develop a flow estimation module without down-sampling processing to handle still scenes in the video. Therefore, the final compensated frame $I_{t+1}^\prime$ is obtained by warping the target frame with the total flow

\noindent
\begin{equation}
I_{t + 1}^\prime= {\cal I}\left\{ {{I_{t + 1}}\left( {\Delta _{{\rm{t}} + 1}^c,\Delta _{{\rm{t}} + 1}^f,\Delta _{{\rm{t}} + 1}^s} \right)} \right\},
\end{equation}

\noindent
where ${\Delta _{{\rm{t}} + 1}^c}$, ${\Delta _{{\rm{t}} + 1}^f}$ and ${\Delta _{{\rm{t}} + 1}^s}$ denote the coarse flow, fine flow and still scenes flow, respectively. Second, we find that motion compensation relies to a large extent on the accuracy of motion estimation. Therefore, the proposed MC module is firstly trained under the supervision of raw frames to get a more accurate motion estimation, then the whole network is jointly fine-tuned based on this MC module.

To verify the effectiveness of our proposed MC module, we present the error maps between two consecutive frames $I_t$ and $I_{t+1}$ in Fig. 2. One can see from Fig. 2 that using the proposed MC operation induces less error in the compensated frame, and our proposed MC method can well eliminate interference in the adjacent frame.

\vspace{-1em}
\subsection{Multi-frame Fusion Module}
The CNN-based temporal information fusion methods have been proposed for various applications, which are mainly classified into early fusion \cite{Early}, slow fusion \cite{Slow} and 3D convolutions \cite{3D}. Early fusion is one of the most straightforward fusion methods, which collapses all temporal information in the first layer. Slow fusion partially merges temporal information in a hierarchical structure and it is slowly fused as information progresses through the network, this fusion method has shown better performance than early fusion for some video applications \cite{Caballero,Slow}. Therefore, we adopt the slow fusion mode as the temporal information fusion method in the SDTS and more details can be found in Fig. 3.

\subsection{Enhancement subnet (ENet)}
The enhancement subnet (ENet) is used to reduce the compression artifacts and enhance the reconstruction effect of target frame in our work. The experimental results in \cite{Hu26} and \cite{IDN} demonstrate that adaptively recalibrating the responses of channel-wise features with coarse-to-fine structure can improve the representation of the network. Therefore, we construct our ENet with a series of coarse-to-fine residual slice blocks (Res$\_$Slice$\_$Block), as shown in Fig. 3. Specifically, only a part of the previous features are delivered to the following modules in each Res$\_$Slice$\_$Block to extract useful information progressively. The local short-path information and the local long-path information are aggregated by concatenation. The slice and concatenation in the Res$\_$Slice$\_$Block are used to control how much the useful information in current state will be reserved and delivered to the next unit. When the weights of both the operators are close to zeros, the information delivered from the previous state will be ignored by the current state. Conversely, more useful information in previous state will be delivered to current state.

\subsection{Training Strategy}
\noindent{\bf{Phase 1}}
The MC module is firstly trained under the supervision of raw frames $I_0^R$ to get more accurate optical flow information. The loss of MC module can be written as
\begin{equation}
{{\cal L}_{ME}} = \sum\limits_{i = - T}^T {{{\left\| {{\cal I}\left( {I_{0 \to i}^R;\Delta _i^R} \right) - I_0^R} \right\|}^2}}.
\end{equation}

\noindent{\bf{Phase 2}} We use Euclidean loss between the reconstructed target frame $I_0^{Rec}$ and the ground truth $I_0^H$ to train the quality enhancement subnet,
\begin{equation}
{{\cal L}_{ENet}} = \sum\limits_{i = - T}^T {{{\left\| {I_0^H - I_0^{Rec(i)}} \right\|}^2}}.
\end{equation}

\noindent{\bf{Phase 3}} We finally fine-tuned the SDTS network
by a joint loss function,
\begin{equation} \label{totalloss}
{\cal L} = {{\cal L}_{ME}} + {\lambda _2}{{\cal L}_{ENet}},
\end{equation}
where ${\lambda _2}$ is the weighting factor that balances the loss terms.

\section{Experiment}
We implement our SDTS framework on TensorFlow platform \cite{TF}. All the experiments are conducted on a PC with Intel Xeon E5 CPU and NVidia GeForce GTX 1080Ti GPU. We conduct all experiments on the same dataset to make a fair comparison between various methods, and all compared methods are retrained over the training dataset according to authors' recommended parameters.

\noindent {\bf{Data Preparation}}
The training and test sequences  in the experiment are compressed in the common test conditions (CTCs) \cite{CTCvvc} by  VVC reference software, VTM3.0, under Low-Delay P (LD) configuration. We specify the Quantization Parameters (QPs) to 32 and 37, respectively. When training the SDTS models, in each video clip, we randomly select the raw frame, its corresponding decoded target frame and the adjacent frames to form the training pairs.

\noindent {\bf{Model Training}}
All the proposed models are trained following the same protocol and share similar hyper-parameters. Filter sizes are set to 3$\times$3, and all non-linearities are rectified linear units except for the output layer, which uses a linear activation.
During training, we use a mini-batch size of 8. To minimize the loss functions of (\ref{totalloss}), ${\lambda _2}$ is empirically set to 0.01, we employ Adam optimizer \cite{Adam} with a start learning rate of 1e-4, decay the learning rate with a power of 10 at the $10^{th}$ epochs, and terminate training at 30 epochs. To save training time, we first train the model at QP 37 from scratch, and the model at QP 32 is fine-tuned from it.

In VVC, the distance between two HQFs that encoded under the LD configuration is normally less than five frames, such a short distance indicates that there exist high correlations among adjacent frames.  As mentioned earlier, the LQFs may be enhanced using the adjacent HQFs. In addition, since the quality fluctuations across compressed video frames under LD configuration, it is difficult to enhance all video frames by utilizing a single model.
In this work, we train a separate model for LQFs and HQFs, respectively, to enhance the quality of VVC compressed video. 
Both the trained models for the quality enhancement of LQFs and HQFs  are proposed by taking advantage of the nearest adjacent HQFs of the corresponding target frame. 

\renewcommand{\arraystretch}{2.2}
\begin{table}[t]
\centering
\fontsize{5.5}{6.5}\selectfont
\caption{Comparisons of different methods on $\Delta$PSNR (dB) over VTM3.0 baseline at QPs 37 and 32}
\label{tab:table1}
\centering
\begin{tabular}{|c|c|c|c|c|c|c|c|}
\cline{1-8}
\multicolumn{1}{|c|}{\multirow{1}{*}{{\bf QP}}}
&\multicolumn{1}{c|}{\multirow{1}{*}{{\bf Class}}}
&\multicolumn{1}{c|}{\multirow{1}{*}{{\bf Seq.}}}
&\multicolumn{1}{c|}{\makecell{\bf VRCNN\\ \cite{Dai14} }}
&\multicolumn{1}{c|}{\makecell{\bf DCAD\\ \cite{Wang15_17Chao}}}
&\multicolumn{1}{c|}{\makecell{\bf MFQE\\ \cite{Yang16} }}
&\multicolumn{1}{c|}{\makecell{\bf SDTS \\ (SF)}}
&\multicolumn{1}{c|}{\makecell{\bf SDTS\\ (MC)}}\\
\cline{1-8}
\multicolumn{1}{|c|}{\multirow{16}{*}{{\bf 37}}}
&\multicolumn{1}{c|}{\multirow{5}{*}{{\bf B}}}
&\multicolumn{1}{c|}{Kimono1}
&\multicolumn{1}{c|}{0.0656}
&\multicolumn{1}{c|}{0.0743}
&\multicolumn{1}{c|}{0.1609}
&\multicolumn{1}{c|}{0.1793}%
&\multicolumn{1}{c|}{0.2687}\\
\cline{3-8}
&\multicolumn{1}{c| }{\multirow{1}{*}{}}
&\multicolumn{1}{c|}{ParkScene}
&\multicolumn{1}{c|}{0.1013}
&\multicolumn{1}{c|}{0.1285}
&\multicolumn{1}{c|}{0.2404}
&\multicolumn{1}{c|}{0.2485}%
&\multicolumn{1}{c|}{0.3664}\\
\cline{3-8}
&\multicolumn{1}{c| }{\multirow{1}{*}{}}
&\multicolumn{1}{c|}{Cactus}
&\multicolumn{1}{c|}{0.0832}
&\multicolumn{1}{c|}{0.0746}
&\multicolumn{1}{c|}{0.1548}
&\multicolumn{1}{c|}{0.2232}%
&\multicolumn{1}{c|}{0.2799}\\
\cline{3-8}
&\multicolumn{1}{c| }{\multirow{1}{*}{}}
&\multicolumn{1}{c|}{BasketballDrive}
&\multicolumn{1}{c|}{0.0452}
&\multicolumn{1}{c|}{0.0711}
&\multicolumn{1}{c|}{-0.0421}
&\multicolumn{1}{c|}{-0.0571}%
&\multicolumn{1}{c|}{-0.0376}\\
\cline{3-8}
&\multicolumn{1}{c| }{\multirow{1}{*}{}}
&\multicolumn{1}{c|}{BQTerrace}
&\multicolumn{1}{c|}{0.1249}
&\multicolumn{1}{c|}{0.1932}
&\multicolumn{1}{c|}{0.1953}
&\multicolumn{1}{c|}{0.2375}%
&\multicolumn{1}{c|}{0.2882}\\
\cline{2-8}
&\multicolumn{1}{c| }{\multirow{4}{*}{{\bf C}}}
&\multicolumn{1}{c|}{RaceHorsesC}
&\multicolumn{1}{c|}{0.0671}
&\multicolumn{1}{c|}{0.0691}
&\multicolumn{1}{c|}{0.0583}
&\multicolumn{1}{c|}{0.0738}%
&\multicolumn{1}{c|}{0.1412}\\
\cline{3-8}
&\multicolumn{1}{c| }{\multirow{1}{*}{}}
&\multicolumn{1}{c|}{BQMall}
&\multicolumn{1}{c|}{0.1029}
&\multicolumn{1}{c|}{0.1232}
&\multicolumn{1}{c|}{0.2155}
&\multicolumn{1}{c|}{0.3396}%
&\multicolumn{1}{c|}{0.3835}\\
\cline{3-8}
&\multicolumn{1}{c| }{\multirow{1}{*}{}}
&\multicolumn{1}{c|}{PartyScene}
&\multicolumn{1}{c|}{0.0634}
&\multicolumn{1}{c|}{0.0582}
&\multicolumn{1}{c|}{0.1206}
&\multicolumn{1}{c|}{0.2600}%
&\multicolumn{1}{c|}{0.3044}\\
\cline{3-8}
&\multicolumn{1}{c| }{\multirow{1}{*}{}}
&\multicolumn{1}{c|}{BasketballDrill}
&\multicolumn{1}{c|}{0.0704}
&\multicolumn{1}{c|}{0.1006}
&\multicolumn{1}{c|}{0.1056}
&\multicolumn{1}{c|}{0.1705}%
&\multicolumn{1}{c|}{0.1984}\\
\cline{2-8}
&\multicolumn{1}{c| }{\multirow{4}{*}{{\bf D}}}
&\multicolumn{1}{c|}{RaceHorses}
&\multicolumn{1}{c|}{0.1407}
&\multicolumn{1}{c|}{0.1763}
&\multicolumn{1}{c|}{0.2192}
&\multicolumn{1}{c|}{0.2071}%
&\multicolumn{1}{c|}{0.3567}\\
\cline{3-8}
&\multicolumn{1}{c| }{\multirow{1}{*}{}}
&\multicolumn{1}{c|}{BQSquare}
&\multicolumn{1}{c|}{0.1893}
&\multicolumn{1}{c|}{0.1947}
&\multicolumn{1}{c|}{0.1895}
&\multicolumn{1}{c|}{0.4081}%
&\multicolumn{1}{c|}{0.4460}\\
\cline{3-8}
&\multicolumn{1}{c| }{\multirow{1}{*}{}}
&\multicolumn{1}{c|}{BlowingBubbles}
&\multicolumn{1}{c|}{0.0833}
&\multicolumn{1}{c|}{0.1206}
&\multicolumn{1}{c|}{0.2371}
&\multicolumn{1}{c|}{0.3031}%
&\multicolumn{1}{c|}{0.3754}\\
\cline{3-8}
&\multicolumn{1}{c| }{\multirow{1}{*}{}}
&\multicolumn{1}{c|}{BasketballPass}
&\multicolumn{1}{c|}{0.1251}
&\multicolumn{1}{c|}{0.1455}
&\multicolumn{1}{c|}{0.3027}
&\multicolumn{1}{c|}{0.3866}%
&\multicolumn{1}{c|}{0.4882}\\
\cline{2-8}
&\multicolumn{1}{c| }{\multirow{3}{*}{{\bf E}}}
&\multicolumn{1}{c|}{FourPeople}
&\multicolumn{1}{c|}{0.2185}
&\multicolumn{1}{c|}{0.2232}
&\multicolumn{1}{c|}{0.2997}
&\multicolumn{1}{c|}{0.4593}%
&\multicolumn{1}{c|}{0.4761}\\
\cline{3-8}
&\multicolumn{1}{c| }{\multirow{1}{*}{}}
&\multicolumn{1}{c|}{Johnny}
&\multicolumn{1}{c|}{0.2029}
&\multicolumn{1}{c|}{0.2105}
&\multicolumn{1}{c|}{0.2906}
&\multicolumn{1}{c|}{0.3957}%
&\multicolumn{1}{c|}{0.4256}\\
\cline{3-8}
&\multicolumn{1}{c| }{\multirow{1}{*}{}}
&\multicolumn{1}{c|}{KristenAndSara}
&\multicolumn{1}{c|}{0.1943}
&\multicolumn{1}{c|}{0.2013}
&\multicolumn{1}{c|}{0.2934}
&\multicolumn{1}{c|}{0.4272}%
&\multicolumn{1}{c|}{0.4577}\\
\cline{2-8}
&\multicolumn{2}{c| }{\multirow{1}{*}{{\bf{Average}}}}
&\multicolumn{1}{c|}{\bf 0.1174}
&\multicolumn{1}{c|}{\bf 0.1353}
&\multicolumn{1}{c|}{\bf 0.1939}
&\multicolumn{1}{c|}{{\bf{0.2664}}}
&\multicolumn{1}{c|}{{{\bf{0.3262}}}}\\
\cline{1-8}
\multicolumn{8}{c}{~}\\[-8pt]
\cline{1-8}
\multicolumn{1}{|c| }{\multirow{1}{*}{{\bf{32}}}}
&\multicolumn{2}{c|}{\bf Average}
&\multicolumn{1}{c|}{\bf 0.1004}
&\multicolumn{1}{c|}{\bf 0.1251}
&\multicolumn{1}{c|}{\bf 0.1974}
&\multicolumn{1}{c|}{{\bf{0.2703}}}
&\multicolumn{1}{c|}{{{\bf{0.2871}}}}\\
\cline{1-8}
\end{tabular}
\end{table}

\subsection{Quantitative Evaluation}
To verify the performance of the proposed SDTS-based method, we evaluate the performance of our SDTS method in terms of $\Delta$PSNR, which measures the PSNR difference between the enhanced and the original compressed frame. We compare our SDTS-based method with some state-of-the-art algorithms, that is, VRCNN \cite{Dai14}, DCAD \cite{Wang15_17Chao} and MFQE \cite{Yang16}. Particularly, VRCNN and DCAD are single-frame based methods, while MFQE is a multi-frame based video quality enhancement method. 
In addition, in order to verify the effect of MC module in the temporal  fusion subnet, we also retrained a model that disables the MC module as a comparison. The temporal fusion component of the comparison model is only  implemented by slow fusion, and the corresponding retrained model is denoted as SDTS$\_$SF. The trained model with MC module is denoted SDTS$\_$MC.

Table \ref{tab:table1} presents the $\Delta$PSNR results of each test sequence at QPs 37 and 32. It can be seen from Table \ref{tab:table1} that our SDTS method outperforms (on average) the other methods. Specifically, the highest $\Delta$PSNR of our SDTS reaches 0.4882dB for MC mode at QP 37, the averaged $\Delta$PSNR gains of our SDTS method is 0.3262dB and 0.2664dB for MC and SF modes, respectively, which are much higher than that of MFQE method (0.1939dB), the state-of-the-art method.
 Besides, when compared with VRCNN and DCAD methods, the SDTS method can even achieve a much higher PSNR gain.
As shown in Table 1, this result has a similar trend at QP 32.
Based on these results, we can conclude that the spatial prior information and multi-frame temporal information play important roles in the quality enhancement for VVC compressed videos.

\vspace{-0.3em}
\subsection{Quality Fluctuation}
We also compare the quality fluctuation of compressed video between various methods. As shown in Fig. 4, we provide the quantified $\Delta$PSNR results for 15 consecutive frames of the test video ``BlowingBubbles". 
Although we only present the results of this test sequence in Fig. 4, we find that all other test sequences have similar results. 
From Fig. 4, one can see that the $\Delta$PSNR curve of our SDTS method is always over the $\Delta$PSNR curves of comparison methods, which indicates that our method can reach a higher $\Delta$PSNR gain for each single frame than the comparison methods. Therefore, our proposed SDTS method is effective to mitigate the quality fluctuation of VVC compressed video, as well as enhancing the quality of  compressed video. 

\begin{figure}[t]
\centering
\includegraphics[width=85mm] {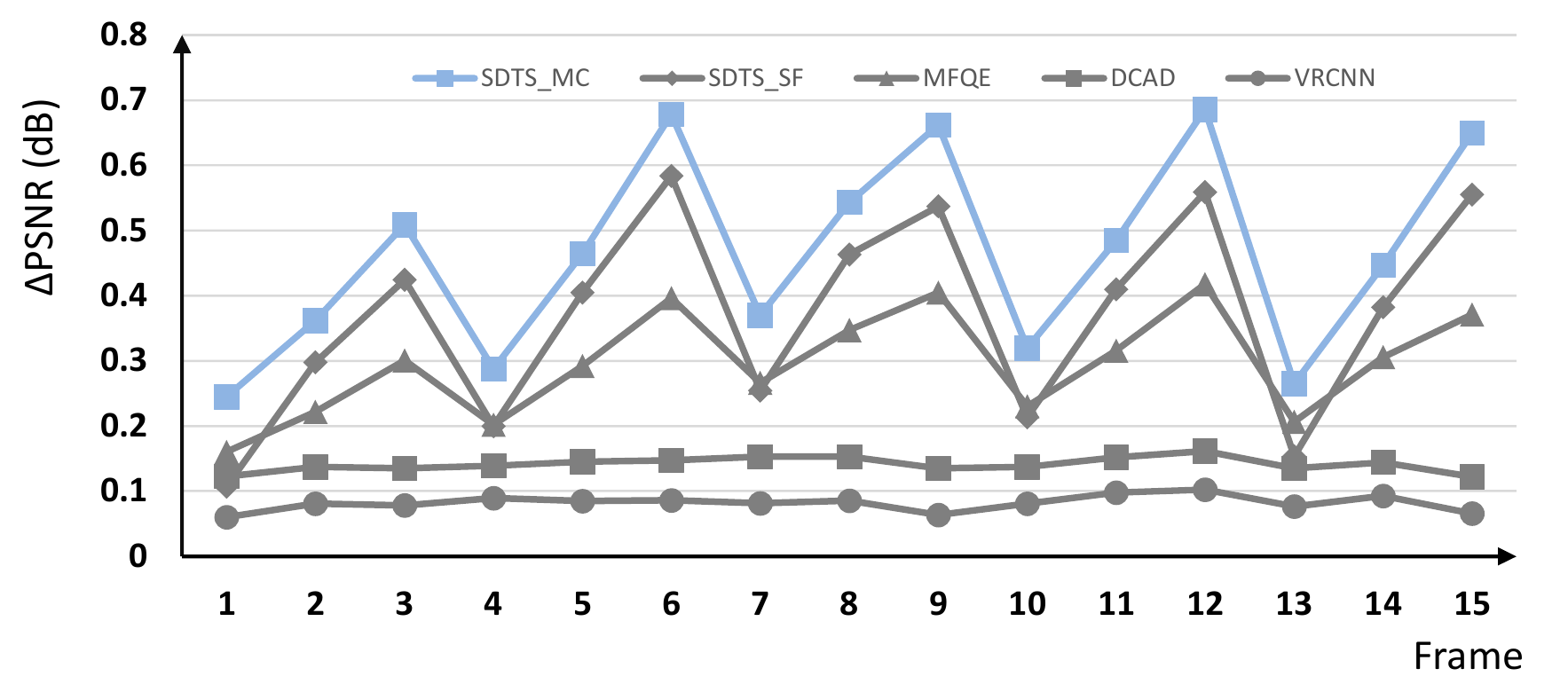}
\caption{Quality fluctuation of ``BlowingBubbles" at QP 37.}
\vspace{-0.3em}
\end{figure}

\vspace{-.2em}
\section{Conclusions}
We proposes a novel CNN-based method to enhance the VVC compressed videos by jointly exploiting spatial details and temporal structure. Our proposed method, i.e. the SDTS-based network, consists of a temporal information fusion subnet and a spatial detail enhancement subnet. The former subnet is utilized to estimate and compensate the temporal motion across frames, and the latter one is employed to enhance the reconstruction quality of the VVC compressed video. Experimental results demonstrate that the proposed SDTS-based method achieves the state-of-the-art performance. 

\vspace{1.5em}
\noindent
\textbf{Acknowledgement}
This work was partially supported by the Applied Basic Research Program of Sichuan Province under Grant 2019YJ0163, by the National Natural Science Foundation of China under Grant 61672134, 61701310, by the Free Exploration Grant for Basic Research of Shenzhen City JCYJ20180305124209486, and by the Fundamental Research Funds for Central Universities of China under Grant ZYGX2016J038.
\bibliographystyle{IEEEbib}
\bibliography{strings,egbib}
\end{document}